\documentclass[runningheads, envcountsame, a4paper]{llncs}
\usepackage{cite}
\usepackage{amsmath,amssymb,amsfonts}
\usepackage{algorithm}
\usepackage{algpseudocode}
\usepackage[pdftex]{graphicx}
\usepackage{textcomp}
\usepackage{xcolor}
\usepackage{stfloats}
\usepackage{multirow}
\usepackage{tabularx}
\usepackage{subcaption}
\usepackage[hyphens]{url}
\usepackage{microtype}
\usepackage[misc]{ifsym}
\algblockdefx{ForAllP}{EndFAP}[1]%
  {\textbf{for all }#1 \textbf{do in parallel}}%
  {\textbf{end for}}
  
\algblockdefx{ForPS}{EndPS}%
  {\textbf{for all parameter server do}}%
  {\textbf{end for}}

\begin{document}
\title{HPSGD: Hierarchical Parallel SGD With\\ Stale Gradients Featuring
		\thanks{This work is supported in part by the National Key Research and Development Program of China under Contract 2017YFB1002201, in part by the National Natural Science Fund for Distinguished Young Scholar under Grant 61625204, and in part by the State Key Program of the National Science Foundation of China under Grant 61836006.
		}
	}
	
\author{Yuhao~Zhou$^{1}$ \and
Qing~Ye$^{2}$ \and
Hailun~Zhang$^{3}$ \and
Jiancheng~Lv~\Letter{}~$^{4}$\thanks{Corresponding author}}

\authorrunning{Zhou et al.}

\institute{College of Computer Science\\
Sichuan University\\
Chengdu, China\\
\email{\{sooptq$^{1}$,tamakokoodaza$^{3}$\}@gmail.com, fuyeking$^{2}$@stu.scu.edu.cn, lvjiancheng$^{4}$@scu.edu.cn}}

\toctitle{HPSGD: Hierarchical Parallel SGD With Stale Gradients Featuring}
\tocauthor{Yuhao~Zhou, Qing~Ye, Hailun~Zhang, Jiancheng~Lv}

\maketitle              

\begin{abstract}
While distributed training significantly speeds up the training process of the deep neural network (DNN), the utilization of the cluster is relatively low due to the time-consuming data synchronizing between workers. To alleviate this problem, a novel Hierarchical Parallel SGD (HPSGD) strategy is proposed based on the observation that the data synchronization phase can be paralleled with the local training phase (i.e., Feed-forward and back-propagation). Furthermore, an improved model updating method is unitized to remedy the introduced stale gradients problem, which commits updates to the replica (i.e., a temporary model that has the same parameters as the global model) and then merges the average changes to the global model. Extensive experiments are conducted to demonstrate that the proposed HPSGD approach substantially boosts the distributed DNN training, reduces the disturbance of the stale gradients and achieves better accuracy in given fixed wall-time.

\end{abstract}

\section{Introduction}
While the synchronous stochastic gradient descent (SSGD) remarkably reduces the training time of the large-scale DNN on the complex dataset by allocating the overall workload to multiple workers, it is additionally required to synchronize local gradients of the workers to keep the convergence of the models~\cite{Dean2012Large}. Hence, the introduced gradient synchronizing phase in the cluster will consume much time, making the acceleration effect of the distributed training non-linear and deteriorating the cluster's scalability. Thus, the communication cost caused by the network I/O and transmission of the synchronization generally becomes the most significant bottleneck of the distributed DNN training with the increasing number of workers and model parameters~\cite{COS}, especially when the communication-to-computation ratio is high (e.g., Gate Recurrent Unit~(GRU)~\cite{cho2014learning}).

Flourish developments have been made to overcome this problem, including batch-size enlarging~\cite{jia2018highly}, periodically synchronizing~\cite{stich2018local} and data compressing~\cite{karimireddy2019error, lin2017deep, yu2018gradiveq, xu2020compressed}. However, although these methods considerably reduce the communication load, many side effects are brought by them to the distributed DNN training process as well, respectively be it the generalization ability degradation~\cite{hoffer2017train}, the added performance-influential hyper-parameter $\gamma$ (i.e., configuration of the interval between synchronizations) or the introduced time-consuming extra phases during training (e.g., sampling, compressing, decompressing, etc.). Moreover, they are all focused on reducing either the worker-to-worker communication rounds or the data transfer size, which limits the results they can achieve since neither the round nor the size can be reduced to $0$.

\begin{figure}[htb]
		\centering
		\includegraphics[width=0.65\linewidth]{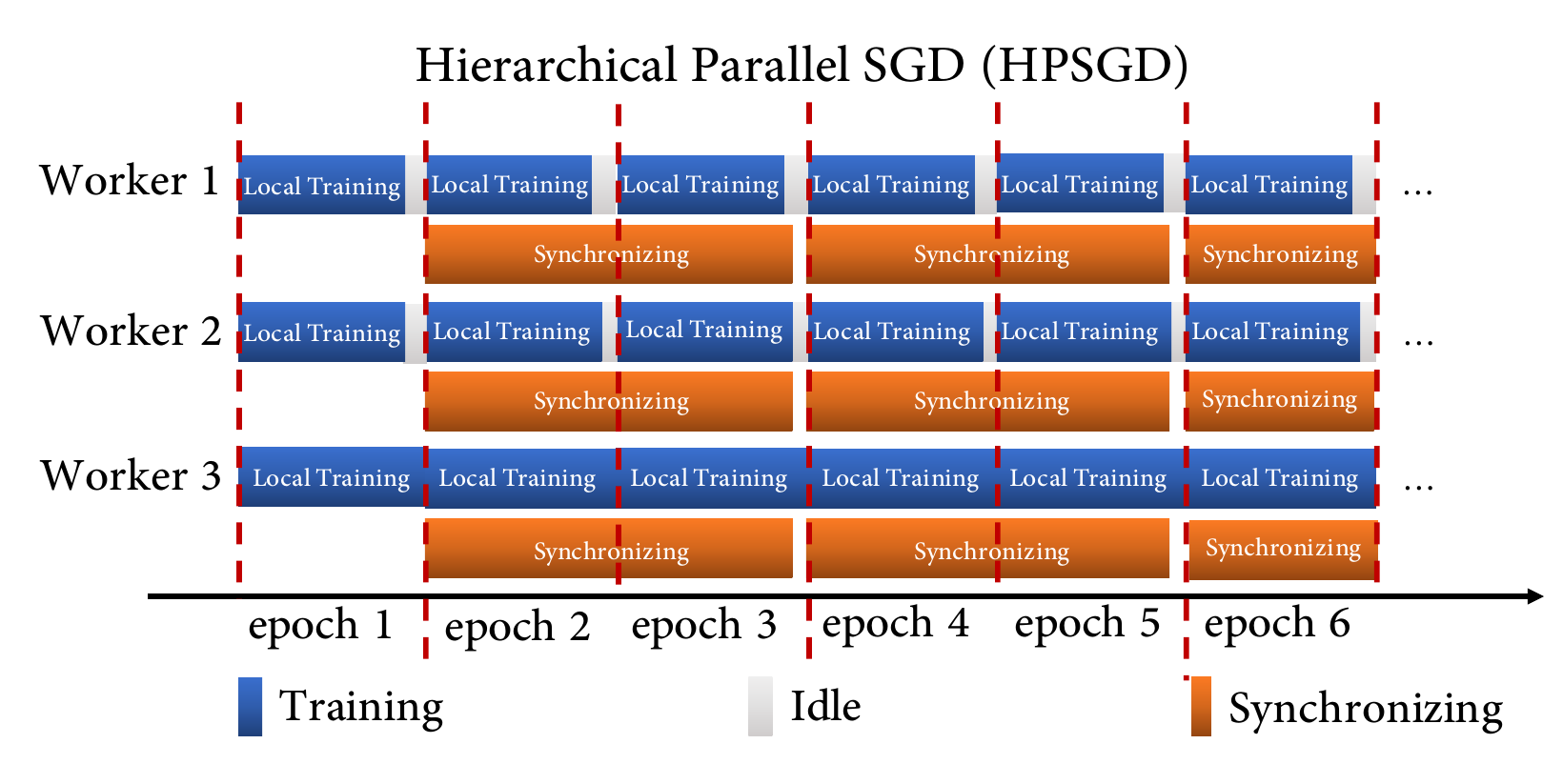}
		\caption{In HPSGD, every worker has two processes. One of them is the local training process doing continuous model training and the other one is the synchronizing process doing continuous data exchanging. These two processes run in parallel.}
		\label{fig:HPSGD-timeline}
\end{figure}

In this paper, we propose Hierarchical Parallel SGD (HPSGD) algorithm that not only fully overlaps the synchronization phase with the local training phase with hierarchical computation but also mitigates the gradients staleness problem and therefore achieves high performances. The desired timeline of HPSGD is illustrated in Fig~\ref{fig:HPSGD-timeline}, which implies that it also ensures synchronous training progresses between workers (i.e., workers start to feed-forward at the same time). The main challenge of all algorithms that separate the local training phase and synchronization phase, including HPSGD, is the gradients staleness problem, meaning the model is updated using stale gradients, which is a detriment to the model convergence. However, Unlike previous literature that tries to counteract stale gradients' effects~\cite{zhang2015staleness}, HPSGD treats these gradients as the features of unknown global optimization surface and thus uses these features to optimize the global training function. In this scenario, the local training phase that overlaps with the synchronization phase helps the global training function collect valuable gradients information and optimize. As a result, the HPSGD algorithm fully utilizes the computational performance, and also maintains model convergence.

The contributions of this paper are summarised as follow:

\begin{itemize}
	\item We entirely overlap the synchronization phase with the local training phase by utilizing hierarchical computation, which significantly boosts the distributed training process.
	\item We utilize an optimized algorithm based on hierarchical computation to address the gradients staleness problem, and therefore improving the training speed, stability, and model accuracy of the distributed DNN training process.
	\item We demonstrate and verify the reliability and effectiveness of HPSGD by applying it to sufficient experiments with various approaches to extensive models. The source code and parameters of all experiments are open-sourced for reproducibility\footnote{\url{https://github.com/Soptq/Hierarchical_Local_SGD}}.
\end{itemize} 
	
The rest of this paper is organized as follows. The literature review is illustrated in Section~\ref{section:two}, where some background information is introduced. In Section~\ref{section:three}, the structure and implementation of the proposed HPSGD algorithm are presented. Then the experimental design and result analysis are detailedly documented in Section~\ref{section:four}. Finally, the conclusions of this paper are drawn in Section~\ref{section:five}.
	
\section{Literature review}
\label{section:two}

\begin{figure}
    \begin{subfigure}{0.49\textwidth}
        \includegraphics[width=\linewidth]{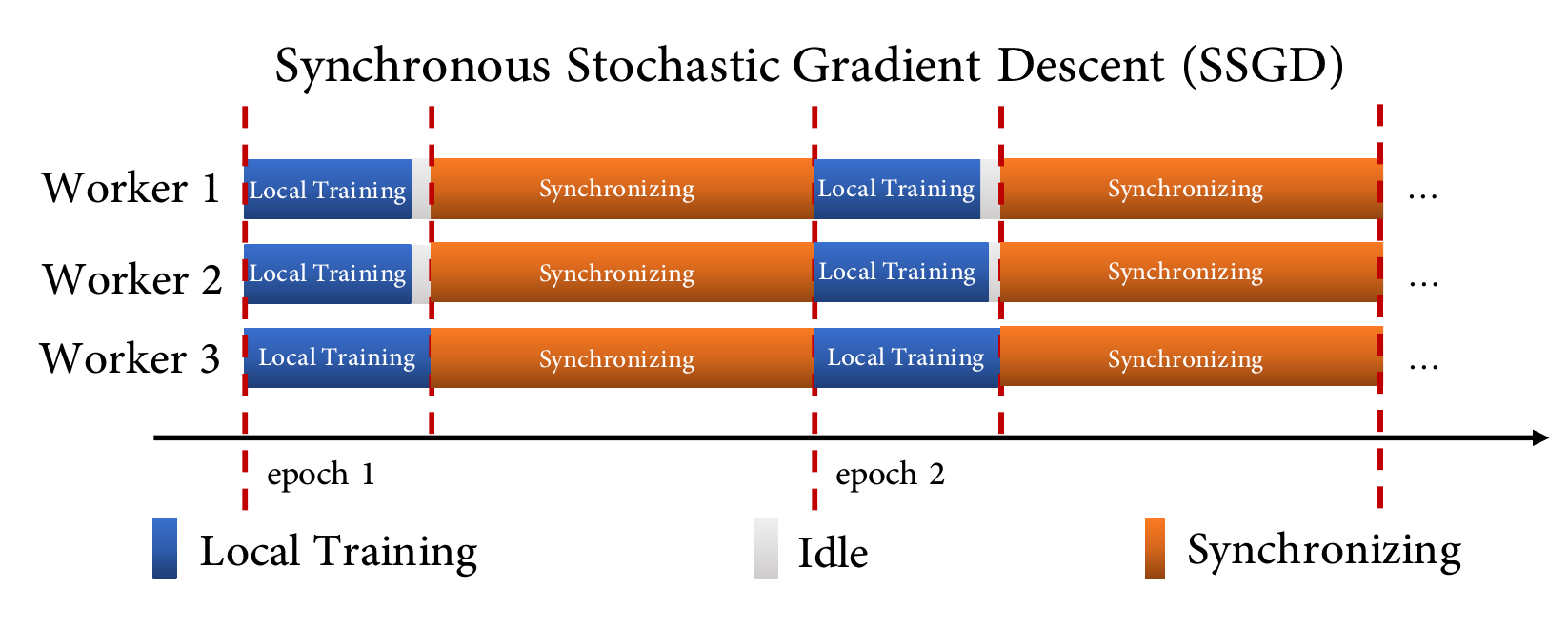}
        \caption{} 
        \label{fig:ssgd-timeline}
    \end{subfigure}%
    \hspace*{\fill}   
    \begin{subfigure}{0.49\textwidth}
        \includegraphics[width=\linewidth]{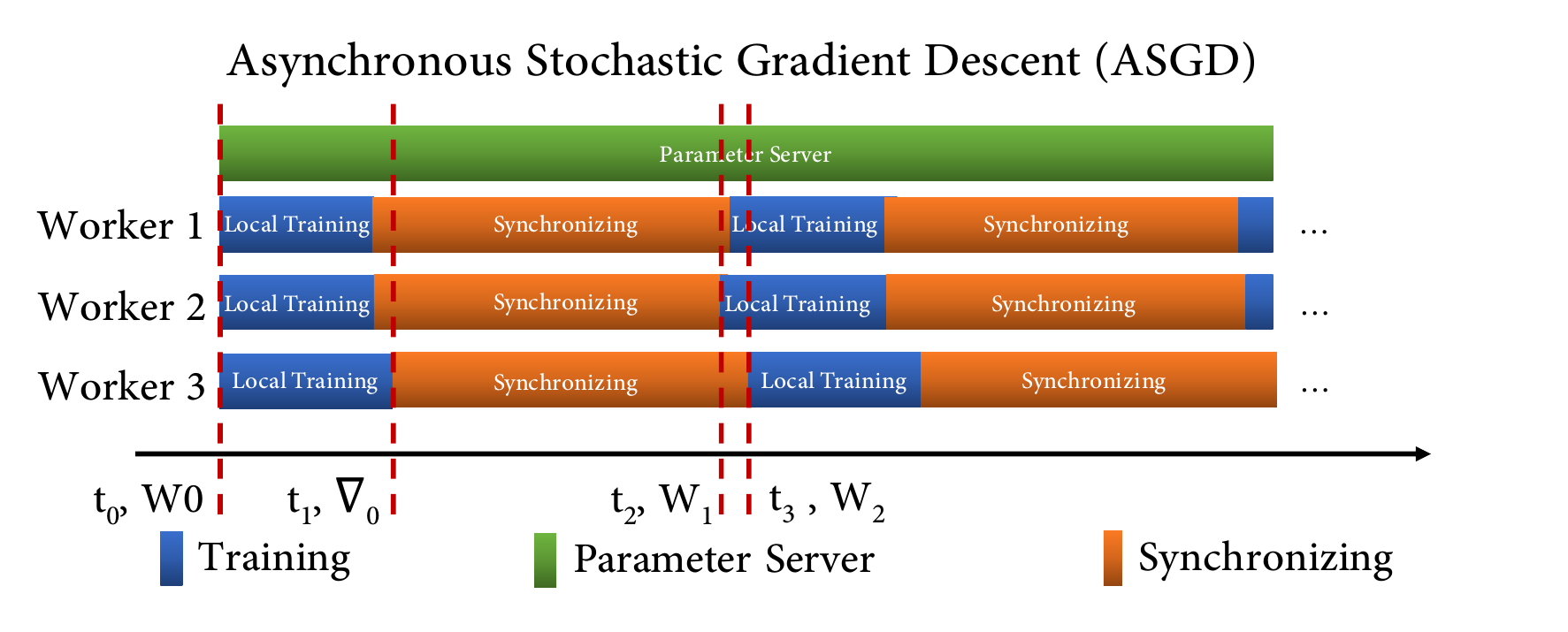}
        \caption{} 
        \label{fig:asgd-timeline}
    \end{subfigure}%
    \caption{(a) A worker waits for other workers to finish local training before starting data exchanging. (b) A worker instantly exchanges data with the parameter server and then steps into the next epoch.}
\end{figure}

\textbf{Synchronous and asynchronous SGD}:
Synchronous SGD (SSGD) is generally a distributed training's model updating strategy that evenly distributes the workload among multiple workers. Then, it updates the model parameters by utilizing SGD algorithm with global gradients aggregated by averaging all local gradients of the different workers. Particularly, the convergence of the model is unaffected with SSGD since it ensures the synchronized gradients are the latest. SSGD can be employed in both centralized\cite{li2014scaling, li2014communication} and decentralized\cite{lian2017can} architectures and the timeline of decentralized SSGD is drawn in Fig~\ref{fig:ssgd-timeline}, where it can be noticed that before starting synchronizing, there is a waiting phase where some workers might have already finished local training and wait for the slower workers to catch up, which leads to a wasted resource. Asynchronous SGD (ASGD) overcomes this problem by allowing workers to work independently. Specifically, fast workers instantly $push$ the calculated local gradients to the parameter servers once they finished training. The timeline of ASGD is shown in Fig~\ref{fig:asgd-timeline}. Although ASGD eliminates the waiting time before synchronizing, it can be utilized only in the centralized architecture, indicating that the cluster is more likely to incur communication overload.

Moreover, as workers are not aware of other workers' status, the gradients staleness problem can be easily triggered. For example, $worker_i$ uses $W_0$ to compute local gradients $\nabla_0$ and synchronizes $\nabla_0$ to parameter servers to start a global model updating operation. However, the global model is updated to $W_1$ during its synchronization phase due to the faster synchronization speed of another worker. Thus $worker_i$ eventually updates the global model $W_1$ to $W_2$ using $\nabla_0$ computed by $W_0$, which will considerably impact the convergence of the model.

It is worth noting that in both centralized and decentralized architectures, synchronizations are processed in the worker's main thread, implying the next epoch's training will be prevented until the current epoch's synchronization phase is completed. Consequently, in both SSGD and ASGD, the processing units of the worker (e.g., CPU, GPU) are idle during the synchronization phase, which is generally a much bigger waste of the worker's performance compared to the waiting phase in SSGD, considering synchronizing usually takes much more time than waiting in practice. 

\textbf{Local SGD}:
Local SGD~\cite{stich2018local} is a well-known algorithm that utilizes periodically model averaging to reduce the number of synchronization rounds .It is capable of achieving good performance both theoretically and practically. Specifically, it introduces a new hyper-parameter $\gamma$ that configures the frequency of the model synchronizing. When synchronizing, workers synchronize the model parameters in place of gradients. However, there are several drawbacks of Local SGD and its variations~\cite{haddadpour2019local, haddadpour2019trading}. 1) Local SGD delivers a relatively slow convergence rate per epoch, and the introduced hyper-parameter $\gamma$ is required to be configured manually to achieve the model's best performance. 2) Although Local SGD reduces the number of synchronization rounds, the computing performance is still idle and not been fully utilized during synchronizing. The pseudo-code of the Local SGD is illustrated in Algorithm~\ref{alg:LSGD}.

\begin{algorithm}[h] 
	\caption{Local SGD} 
	\label{alg:LSGD} 
	\begin{algorithmic}[1] 
		\State Initialization: Cluster size $n$. Learning rate $\mu \geq 0$. Max training $epoch$. Local gradient $\widehat{\nabla}^{e}$. Synchronous period $\gamma$;
		\ForAllP {$i \in {1,...,n}$}
			\For {$e \in {1,...,epoch}$}
				\State Update local model: $w_i^{e+1} = w_i^{e}- \mu\widehat{\nabla}^{e}$;
				\If {$e~mod~\gamma == 0 $}
					\State Average model: $w_i^{e+1} = AllReduce(w_i^{e+1})$;
				\EndIf
			\EndFor
	    \EndFAP
	\end{algorithmic} 
\end{algorithm}
	
\section{Methodology}
\label{section:three}
In this section, we present the implementation of the proposed HPSGD detailedly, which includes: 1) Spawning two process $P_s$ and $P_t$ to Perform data synchronizing and local training, respectively. 2) Applying a model updating algorithm to alleviate the gradients staleness problem.

\subsection{Implementation of hierarchical computation}
Hierarchical computation enables the synchronization phase to be fully overlapped with the local training phase and is achieved by spawning a dedicated process $P_s$ responsible for data synchronizing. $P_t$ and $P_s$ are located in the same worker and they share the same rank in the distributed system. These two processes connect and communicate via shared memory, and there are typically the following variables that need to be shared.

\begin{itemize}
    \item \textbf{$status$}: The variable that indicates the status of $P_s$. It has two states: $synchronizing$ and $idling$.
    \item \textbf{$replica$}: The replica of the latest global model, which will be detailedly discussed in Section~\ref{gradienst_utilization}.
    \item \textbf{$\nabla_{i}^{a}$}: The $i$-th worker's accumulated gradients when workers are performing local training.
    \item \textbf{$\widehat{\nabla}_{i}^{e}$}: The local gradients calculated by $i$-th worker at epoch $e$.
    \item \textbf{$counter$}: The integer that represents how many times has $P_t$ trained locally.
\end{itemize}

specifically, when $status$ is $synchronizing$, $P_t$ will firstly make a replica of the global model if the $counter$ equals $0$, then it will accumulated the calculated gradients to the $\nabla_{i}^{a}$, updating $replica$ and finally increasing the $counter$ by $1$. On the other hand, $P_t$ will activate $P_s$ to start to synchronize and mark the $status$ as $synchronizing$ when the $status$ is $idling$, and $P_s$ will firstly update the global model using the global gradients which was synchronized in the last time, and then $AllReduce$ing the $\nabla_{i}^{a}$, resetting $counter$ to $0$, and finally marking the $status$ as $idling$. The workflow of the HPSGD algorithm is demonstrated as algorithm \ref{alg:HPSGD-hierarchical}

\begin{algorithm}[h] 
	\caption{Hierarchical Parallel SGD Algorithm} 
	\label{alg:HPSGD-hierarchical} 
	\begin{algorithmic}[1] 
		\State Initialization: Cluster size $n$. Learning rate $\mu \geq 0$. Max training $epoch$;
		\ForAllP {$i \in {0,...,n-1}$}
			\For {$e \in {0,...,{epoch-1}}$}
				\If {$state == synchronizing$}
					\If {$counter == 0$}
						\State Make replica: $r_i = clone(w_i^e)$;
					\EndIf
	        		\State Update replica: $r_i = r_i- \mu\widehat{\nabla}_{i}^{e}$;
	        		\State Accumulate gradients: $\nabla_{i}^{a} = \nabla_{i}^{a} + \widehat{\nabla}_{i}^{e}$;
	        		\State Increase counter: $counter = counter + 1$;
	        	\Else
	        		\State Accumulate gradients: $\nabla_{i}^{a} = \nabla_{i}^{a} + \widehat{\nabla}_{i}^{e}$
	        		\State Mark $status$: $status = synchronizing$;
	        		\State Instruct $P_s$: $StartSync(P_s)$;
	        		\label{HPSGD-instrut}
	        	\EndIf
	        \EndFor
	    \EndFAP
	\end{algorithmic} 
\end{algorithm}

In step \ref{HPSGD-instrut}, $StartSync$ is a function to activate $P_s$ to start to synchronize. It is an operation processed in $P_s$, which does not block the training process in $P_t$, so that $P_t$ can keep training instead of idling. Furthermore, $StartSync$ function also reduces the gradients staleness effects and will be explained in detail below.

\subsection{Gradients utilization}
\label{gradienst_utilization}
	
\begin{figure}[h]
	\centering
	\includegraphics[width=\linewidth]{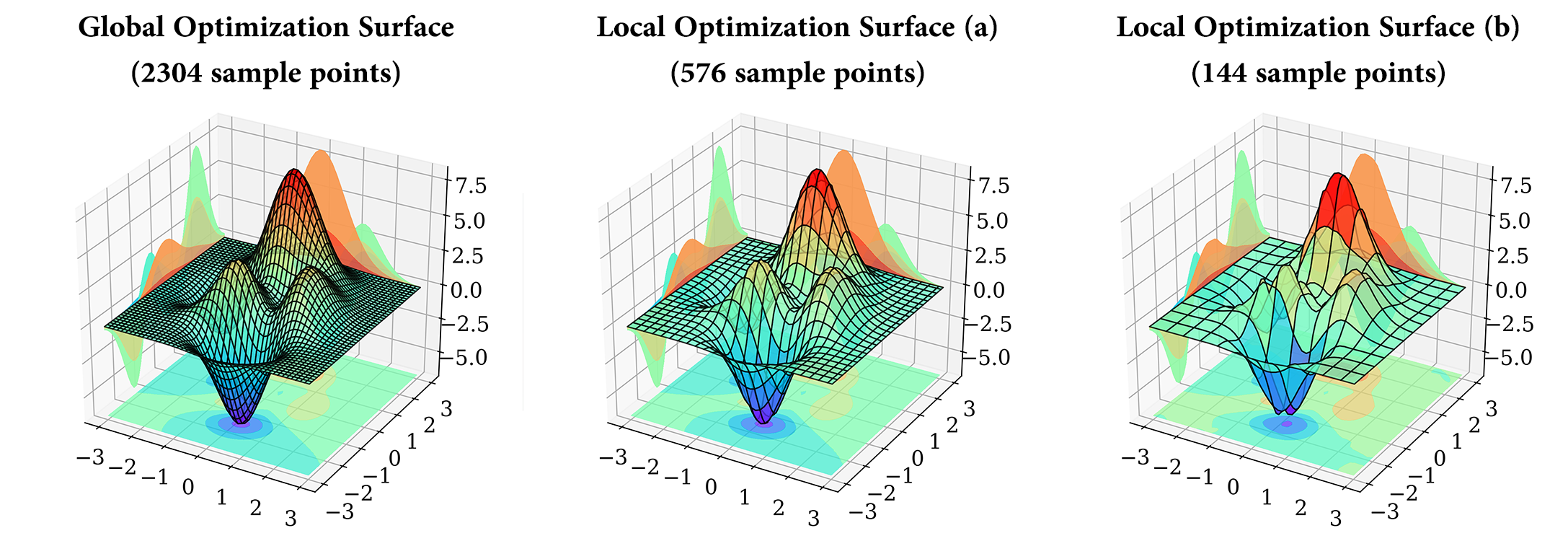}
	\caption{Intersections on the surface represent sample points. The local optimization surface with massive reduced number of sampling points still retains similar features as the global optimization surface, including the coordinates of the extremums, the variations of partial derivatives over an interval, etc. }
	\label{fig:optimization_surface}
\end{figure}	

HPSGD considers stale gradients advanced instead of stale and lets workers make replicas of the current global model when starting to synchronize. When performing local training, calculated $\widehat{\nabla}_{i}^{e}$ will be applied to workers' respective replicas and be added to $\nabla_{i}^{a}$. Finally, when the synchronization completes, accumulated $\nabla_{i}^{a}$ during the local training will be committed to the previous global model. The key concept of HPSGD's model updating algorithm is that with the increase of the dataset size, the local optimization surface modeled by a worker with a sub-dataset becomes more similar to the global optimization surface. Therefore, the local gradients of different workers on its own sub-dataset can be utilized to help optimize the global training function, which is illustrated in Fig~\ref{fig:optimization_surface}. The HPSGD's model updating function can be formulated as:

\begin{equation}
\label{equ:hpsgd}
	W_i^{e_2+1} = W_i^{e_1} - \mu\frac{\sum_{i=0}^{n}\sum_{e=e_1}^{e_2}\widehat{\nabla}_{i}^{e}}{n}
\end{equation}

Where $W_e$ denotes the model parameters at epoch $e$, $n$ refers to the number of workers in the distributed system and $e_1$, $e_2$ denote the epoch range of local training. The equation can be further simplified to:

\begin{equation}
\label{equ:hpsgd-simp}
	W_{i}^{e_1} - \mu\frac{\sum_{i=0}^{n}\sum_{e=e_1}^{e_2}\widehat{\nabla}_{i}^{e}}{n} = \frac{\sum_{i=0}^{n}W_{i}^{e_2}}{n}
\end{equation}

Which suggests that HPSGD is generally a Local SGD's deformation in the form of gradients, thus the convergence of HPSGD can be proved by \cite{yu2019parallel, stich2018local}. Although the formula is essentially the same for HPSGD and Local SGD, HPSGD focuses on achieving a lock-free and highly paralleled model updating algorithm with minimal influence from gradients staleness problem while Local SGD specializes in synchronization rounds reducing. In addition, since the synchronization phase is overlapped with the local training phase, it is deemed impossible to synchronize and update model parameters at the same time, because the model parameters would then be written simultaneously by $P_s$ and $P_t$ at the next epoch. Thus the simplified model updating function Equ~\ref{equ:hpsgd-simp} can not be used in the real scenario and therefore the gradients are synchronized between workers instead of model parameters to update the model with Equ~\ref{equ:hpsgd} as gradients are generally intermediate data in the model update process and do not need to be persisted. Furthermore, HPSGD eliminates the hyper-parameter $\gamma$. Instead, it continuously performs synchronization in another process (i.e, performs synchronizing whenever possible), making the synchronization phase highly flexible and bringing two benefits to the distributed DNN training process: 1) Improved robustness. Methods like Local SGD fix $\gamma$, assuming the synchronization time is stable throughout the training process, which is not practical in real scenarios where exceptions are unpredictable. 2) the maximal number of synchronizations, which improves the convergence rate and stability of the distributed DNN training process.

For example, assume the local training time is $t_{train}$ seconds and synchronization time is $t_{sync}$ seconds where $t_{sync} = k \cdot t_{train}$. In Local SGD, a complete loop that contains $\gamma$ local trainings and one data synchronizing takes $((\gamma + 1) \cdot t_{train} + t_{sync})$ seconds. For the same period of time, HPSGD can synchronize for averagely $(\frac{\gamma + 1}{k} + 1)$ times and train locally for $(\gamma + k + 1)$ times, suggesting that in a given fixed time, HPSGD could sample more features in optimization surfaces and perform data synchronizing more times. Consequently, HPSGD eliminates the hyper-parameter $\gamma$, while making the global model iterate faster, sample more features and achieve better accuracy. The pseudo-code of these behaviors is presented in Algorithm \ref{alg:HPSGD-synchronizing}.

\begin{algorithm}[h] 
	\caption{$P_s$'s behavior} 
	\label{alg:HPSGD-synchronizing} 
	\begin{algorithmic}[1] 
		\If {$e-counter > 0$}
		\label{DPG:update-condition}
			\State Update global model: $w_i^{e+1} = w_i^{e - counter}- \mu\widehat{\nabla}_{i}^{e - counter}$
		\EndIf
		\State $AllReduce$ gradients: $\widehat{\nabla}_{i}^{e} = AllReduce({\nabla_{i}^{a}})$
		\State Reset the counter: $counter = 0$
		\State Mark the $status$: $status = idle$
	\end{algorithmic} 
\end{algorithm}

As the pseudo-code has shown, $P_s$ $AllReduce$s the gradients and will perform global model updating with these synchronized gradients at the next synchronization's beginning. Thus, step \ref{DPG:update-condition} is used for ensuring that the synchronized gradients exist when performing the first global model updating, since the first epoch is almost certainly used for local training instead of synchronizing in practice. Consequently, the global model updating operation is always $1$ synchronization delayed compared to the corresponding $AllReduce$ operation due to the above updating strategy of $P_s$.

\section{Experiments}
\label{section:four}

\subsection{Experimental setup}
\textbf{Hardwares}: An Nvidia DGX-Station is employed to set up the environment of the experiments with 4 Nvidia Tesla V100 32G GPU and Intel(R) Xeon(R) CPU E5-2698 v4 @ 2.20GHz.

\textbf{Softwares}: All experiments are done in an nvidia-docker environment with CUDA 9.0.176. Pytorch 1.6.0\footnote{\url{https://pytorch.org}} is utilized to simulate the distributed training process of the cluster by spawning multiple processes, whereas each stands for an individual worker.

\textbf{Methods}: SSGD, HPSGD, Local SGD~\cite{stich2018local}, and purely offline training (PSGD, no communication between workers during training). PSGD will only be presented in the convergence rate comparison since it is utilized to only give a reference of fast training speed.

\textbf{Models}: ResNet~\cite{he2016deep}, DenseNet~\cite{huang1608densely}, MobileNet~\cite{sandler2018mobilenetv2} and GoogLeNet~\cite{szegedy2015going}.

\textbf{Datasets}: Cifar-10~\cite{krizhevsky2009learning} dataset, which consists of $60,000$ $32 \times 32$ images in total with both RGB channels.

\textbf{Other settings}: Learning rate: $0.01$, batch size: $128$, $\gamma$ of Local SGD: $8$, epoch size $100$, loss function: cross entropy, optimizer: SGD.

\subsection{Experiment Design and analysis}

\begin{figure}
    \begin{subfigure}{0.49\textwidth}
        \includegraphics[width=\linewidth]{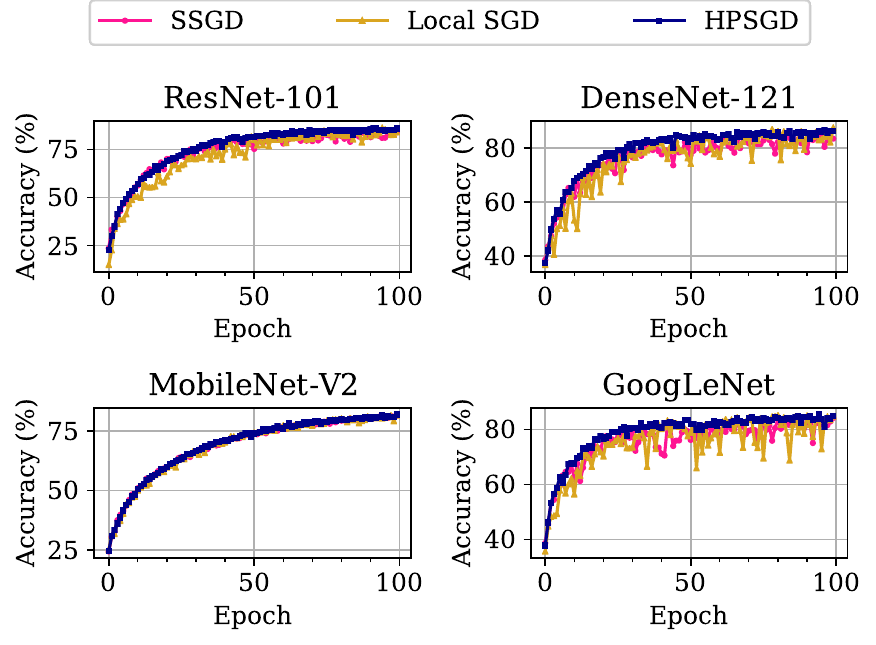}
        \caption{} 
        \label{fig:all-acc}
    \end{subfigure}%
    \hspace*{\fill}   
    \begin{subfigure}{0.49\textwidth}
        \includegraphics[width=\linewidth]{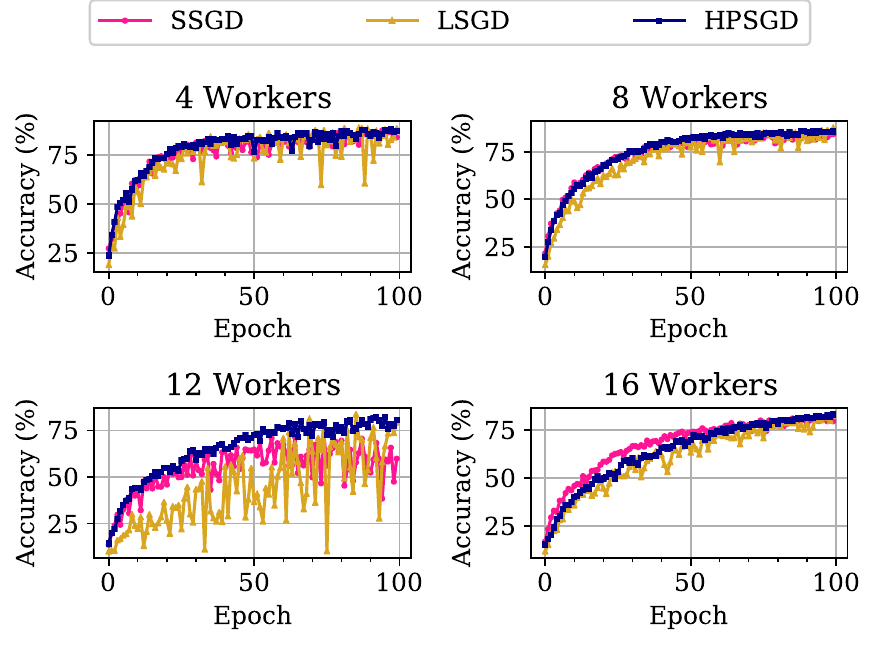}
        \caption{} 
        \label{fig:resnet-scale-acc}
    \end{subfigure}%
    \caption{(a) Accuracy comparisons of various models each with different methods. (b) The accuracy comparison with 4 workers, 8 workers, 12 workers and 16 workers.}
\end{figure}

\textbf{Convergence and training loss}: Various models have been trained with different methods in order to verify the convergence of HPSGD. Accuracy curve of 4 workers with different models and of different cluster size with the ResNet-101 model is presented in Figs~\ref{fig:all-acc} and \ref{fig:resnet-scale-acc}, respectively.

As shown in Fig~\ref{fig:all-acc}, the reached accuracy of HPSGD at epoch 100 is generally identical to the SSGD's among all experiments, suggesting HPSGD maintains the convergence of the model by utilizing local gradients to help global training function optimize. Moreover, the accuracy curve of HPSGD is significantly smoother than other methods, especially in GoogLeNet. This phenomenon is caused by the fact that the global model in HPSGD is updated by gradients that are repeatedly sampled on the sub-dataset between synchronizations, which could considerably reduce the instability of mini-batch SGD.

Furthermore, in some cases, the accuracy of HPSGD even outperforms SSGD (e.g., DenseNet-121). We believe it is mainly due to the HPSGD algorithm's characters that it is less likely to be trapped in a local optimum. Notably, as HPSGD generally lets workers independently compute their solutions and lastly applies them to the global model, it can be seen as the global model is simultaneously optimized toward multiple directions in the global optimization surface. Thus, the probability of multiple models simultaneously trapping in their local optimums is significantly reduced compared to a single model updating toward one direction.

\begin{figure}
    \begin{subfigure}{0.45\textwidth}
        \includegraphics[width=\linewidth]{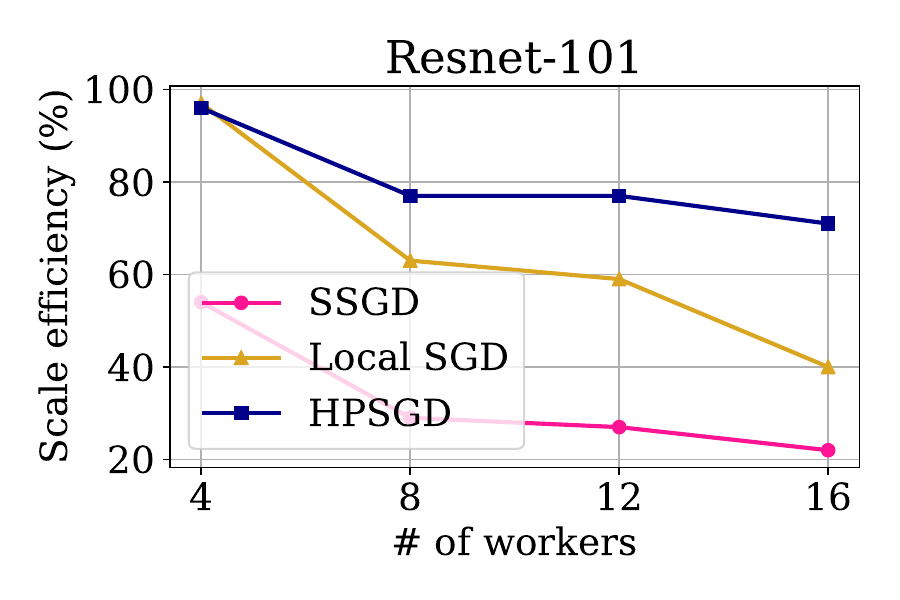}
        \caption{} 
        \label{fig:all-scale}
    \end{subfigure}%
    \hspace*{\fill}   
    \begin{subfigure}{0.45\textwidth}
        \includegraphics[width=\linewidth]{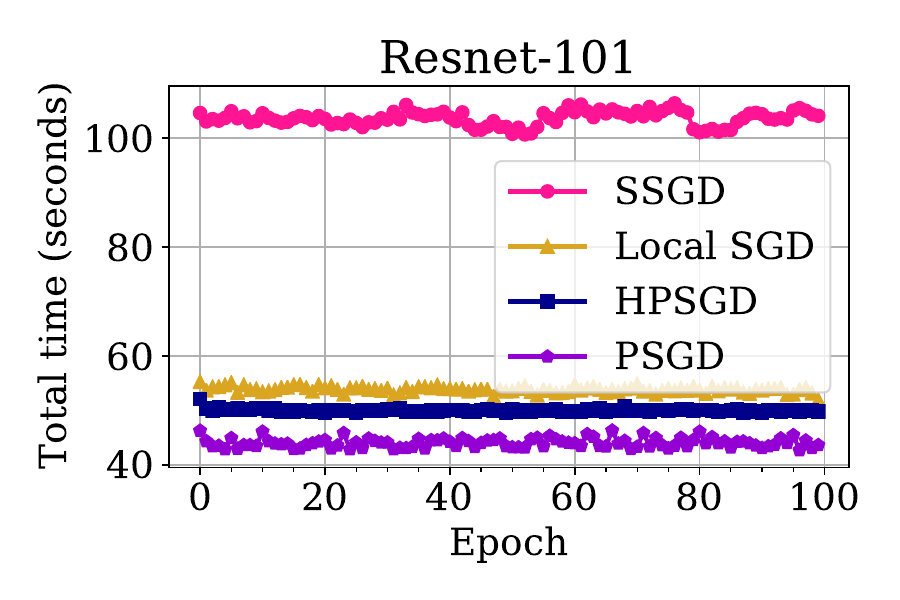}
        \caption{} 
        \label{fig:resnet-realtime}
    \end{subfigure}%
    \caption{(a) The scale efficiency of different cluster size with ResNet-101 model. (b) The total time of different methods training Cifar-10 with ResNet-101.}
\end{figure}

\textbf{Scale efficiency}: The scale efficiency of different methods with the ResNet-101 is shown in Fig~\ref{fig:all-scale}. As the figure demonstrated, both HPSGD and Local SGD have a much larger scale efficiency than SSGD when the number of workers is relatively small with significantly reduced network traffic. However, with the cluster size increase, Local SGD's scale efficiency is drastically decreased, indicating the communication jam is triggered, and its synchronization interval $\gamma$ needs to be larger. However, enlarging the $\gamma$ could lead to a slower convergence rate. Meanwhile, the impacts on HPSGD are considerably smaller than other methods when the number of workers increases. Specifically, HPSGD obtains the same performance compared to Local SGD and 133\% more performances compared to SSGD when four workers participated in the training. When there are 16 workers, HPSGD obtains 75\% more performance compared to Local SGD and 250\% more performance compared to SSGD. This is mainly because, theoretically, in HPSGD, there is no extra synchronizing time at all during distributed DNN training procedure as it is entirely overlapped with the local training phase. Thus, the main reason for the decreasing scale efficiency and the performance loss for HPSGD is the increasing time spent in the waiting phase, which is caused by different computational performance of workers (gray part in Fig~ \ref{fig:HPSGD-timeline}). This will be left as our future optimization direction.

\begin{figure}
    \begin{subfigure}{0.32\textwidth}
        \includegraphics[width=\linewidth]{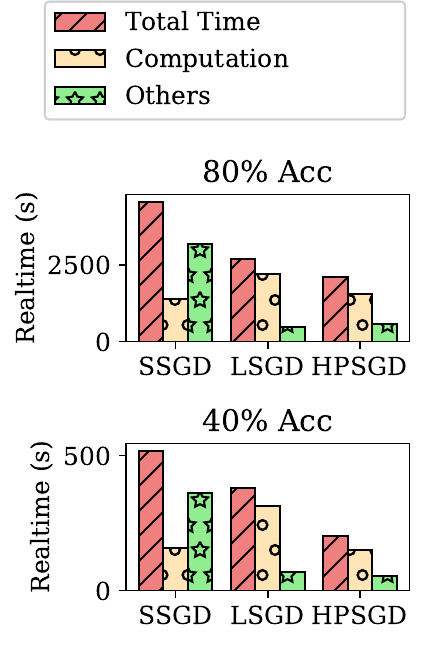}
        \caption{} 
        \label{fig:resnet-realtime-breakdown}
    \end{subfigure}%
    \hspace*{\fill}   
    \begin{subfigure}{0.63\textwidth}
        \includegraphics[width=\linewidth]{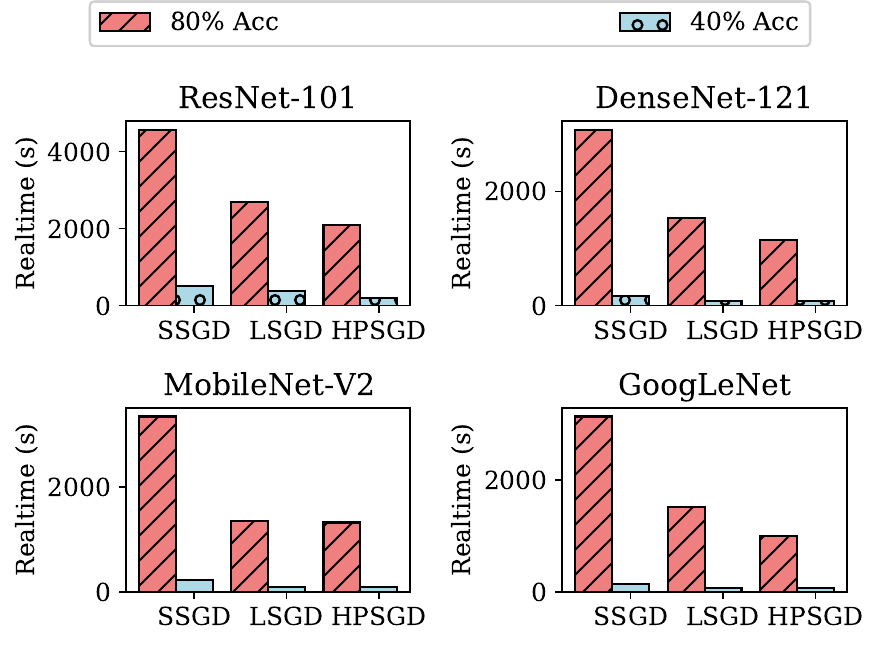}
        \caption{} 
        \label{fig:all-realtime}
    \end{subfigure}%
    \caption{a) Upper: Time breakdown for reaching 80\% accuracy. Lower: Time breakdown for reaching 40\% accuracy. b) Total training time used to reach 80\% and 40\% accuracy on different models with different methods.}
\end{figure}

\textbf{Convergence rate}: Fig~\ref{fig:resnet-realtime} illustrates the cost time for each epoch with different methods. Here LSGD refers to Local SGD due to tight space. PSGD serves as the lower bound of the cost time of the distributed DNN training. The time difference between HPSGD and PSGD is mainly due to the impact of limited CPU time and workers' performance difference. The breakdown of the total training time is presented in Fig~\ref{fig:resnet-realtime-breakdown}. It can be shown that the computation time of SSGD and HPSGD is roughly the same when reaching either 80\% accuracy or 40\% accuracy, suggesting that while HPSGD shares the same converge rate as SSGD, it drastically reduce the non-computation-related time and thereby boosting the distributed training process. On the other hand, although the total time of Local SGD is shorter than SSGD when reaching either 80\% accuracy or 40\% accuracy, the computation time is relatively longer, indicating that the converge rate of Local SGD is lower than the SSGD and HPSGD. This phenomenon matches and verifies the explanation in Section~\ref{gradienst_utilization}. To avoid chance, we performed more experiments on total training time (wall time) of four different models with the same configuration, which is illustrated in Fig~\ref{fig:all-realtime}

\section{Conclusions and future work}
\label{section:five}
In this paper, we propose a novel Hierarchical Parallel SGD (HPSGD) algorithm that firstly overlaps the time-consuming synchronization phase with the local training phase by deploying hierarchical computation across two processes, which significantly boosts the distributed training. Then it alleviates the stale gradients problem by utilizing the sub-gradients calculated by different workers to help global model update. Detailedly, workers perform training on a replica of the global model independently, recording local gradients and lastly committing these gradients to the global model. In such circumstances, the sub-gradients of different workers are not stale but advanced and can be taken advantage of. Extensive experiments and comparisons turn out that the performance of HPSGD surpasses SSGD and Local SGD, which actively verifies its effectiveness and high efficiency. However, although HPSGD drastically drops the synchronization time of the distributed training process, the waiting phase remains, which is caused by workers' imbalanced performances. In future work, we would like to investigate methods capable of reducing such waiting costs and thereby further improving the scalability of the cluster in distributed training.

\bibliographystyle{splncs04}
\bibliography{refs}

\end{document}